\documentclass[runningheads]{llncs}

\usepackage[utf8]{inputenc}
\usepackage{graphicx}
\usepackage{subcaption}
\usepackage{cleveref}
\usepackage{enumitem}
\usepackage{booktabs}
\usepackage{gensymb}
\usepackage[font=small]{caption}
\usepackage{cite}
\usepackage{todonotes}
\usepackage{tabularx, colortbl}
\usepackage{multirow}
\usepackage{bm}
\usepackage{siunitx}
\usepackage{verbatim}
\usepackage[square,numbers]{natbib}

\usepackage{balance}

\usepackage{tikz}
\usetikzlibrary{external}
\usepackage{pgfplots}
\pgfplotsset{compat=1.9}
\usepgfplotslibrary{statistics}

\usepackage{titlesec}
\titlespacing{\section}{5pt}{*0.7}{*0.7}
\titlespacing{\subsection}{10pt}{*0.6}{*0.6}
\usepackage{enumitem}
\setlist[itemize]{noitemsep, topsep=0pt}
\setlist[enumerate]{noitemsep, topsep=0pt}
\renewcommand{\arraystretch}{0.9}

\title{AR Point\&Click: An Interface for Setting Robot Navigation Goals}

\author{Morris Gu\inst{1},  Elizabeth Croft\inst{2} and Akansel Cosgun\inst{3}}
\institute{{Monash University, Australia}\and{University of Victoria, Canada}\and{Deakin University, Australia}}
\begin{document}
\maketitle

\begin{abstract}
This paper considers the problem of designating navigation goal locations for interactive mobile robots. We investigate a point-and-click interface, implemented with an Augmented Reality (AR) headset. The cameras on the AR headset are used to detect natural pointing gestures performed by the user. The selected goal is visualized through the AR headset, allowing the users to adjust the goal location if desired. We conduct a user study in which participants set consecutive navigation goals for the robot using three different interfaces: AR Point\&Click, Person Following and Tablet (birdeye map view). Results show that the proposed AR Point\&Click interface improved the perceived accuracy, efficiency and reduced mental load compared to the baseline tablet interface, and it performed on-par to the Person Following method. These results show that the AR Point\&Click is a feasible interaction model for setting navigation goals.
\end{abstract}
  
\section{Introduction}

The service robot sector is continuing to grow at a strong pace, with an expected 22.6\% cumulative average growth rate from 2020 to 2025~\cite{service_robots}. As these robots move from industrial applications to adoption in public facing and residential use, they are increasingly being run by non-expert users. Interactive robots are typically designed to take commands from users in response to changing demands, as opposed to pre-programmed industrial robots operating in separated assembly lines. As such, there is a demand for user-friendly methods that allow users to set goals for the robots in intuitive ways. For interactive mobile robots, allowing users to set navigation goals is a problem of broad interest.


Several methods of setting navigation goals already exist. One of the most common methods is setting goals from a birdseye map of the environment. A prominent example of this is the RViz tool within Robot Operating System (ROS) which is popular among the robotics developer community. Another common method is sending the robot to semantically meaningful locations, such as to rooms or labeled areas~\cite{cosgun_context-aware_2018}. Other interactive goal setting methods involve co-located direct driving \cite{cosgun2015human}, or letting the robot follow the user \cite{cosgun_autonomous_2013, scales_studying_2020}. In human-to-human interaction, a common method is using deictic (pointing) gestures.

Deictic gestures are also commonly used for human-robot interaction. Some approaches use pointing devices to provide goals to robots. In implementations by Kemp~\cite{kemp2008point} and ~Gualtieri\cite{gualtieri2017open}, a point and click and gesture is implemented using a laser pointer to select the goal objects for fetch-and-carry tasks. Kemp~\cite{kemp2008point} suggest that point and click interfaces are simple and have a diverse set of applications. This concept is extended by Nguyen~\cite{nguyen2008clickable} to driving tasks, although users can only select from a set of pre-defined discrete navigation targets. Sprute~\cite{sprute2019far} proposes a system in which laser pointers can be used to define a virtual barrier for navigation. Chen~\cite{chen2013context} uses the head orientation of the Augmented Reality headset to choose between various target control devices. Alternatively, some designs use natural hand and/or arm gestures for setting robot goals, for instance, for selecting object goals \cite{cosgun2015did} or to point into space for specifying region goal \cite{hato2010pointing}. Our work also employs natural deictic gestures for selecting robot goals.
\begin{figure}[t]
    \centering
    \includegraphics[trim= 5 60 20 17.5, clip,width=0.76\linewidth]{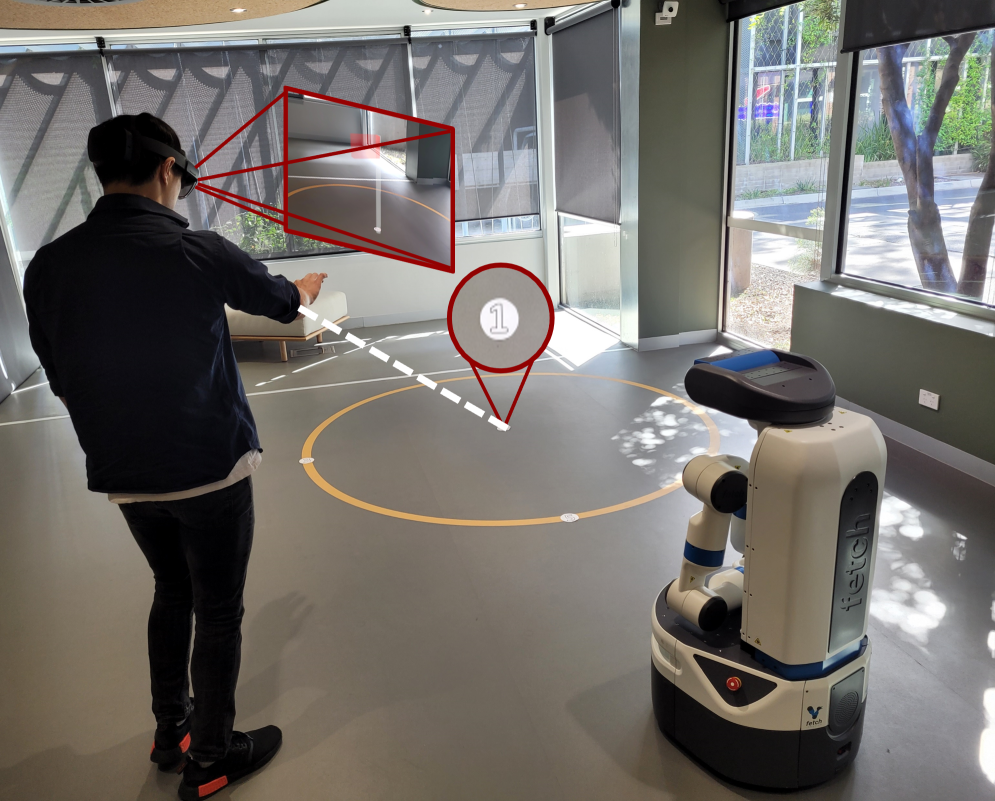}
    \caption{We propose the \textbf{AR Point\&Click} interface which allows users to perform a pointing gesture, followed by a pinching gesture with the fingers to assign a navigation goal to a mobile robot. The chosen goal is visualized through the AR headset as a virtual golf flag pole, which allows the user to view and adjust the goal on the fly.} 
    \vspace{-0.1cm}
    \label{fig:intro}
\end{figure}

Augmented Reality (AR) technology has been growing in popularity for human-robot interaction \cite{makhataeva2020augmented}, with applications ranging from visualizing robots behind walls \cite{gu_seeing_2021}, visualizing the robot's intent for object handovers \cite{newbury_visualizing_2021} or for drones \cite{walker_communicating_2018}, and human-robot cooperative search \cite{reardon_come_2018}. Our previous works from Hoang~\cite{hoang_virtual_2021} and Waymouth~\cite{waymouth_demonstrating_2021} demonstrate choosing goal points for interactive manipulation tasks using natural hand gestures. Kousi~\cite{kousi_enabling_2019} demonstrates an AR application to set navigation goals and teleoperate a virtual robot. 

In this work, we investigate the usability of an AR point and click interface that was initially demonstrated as a proof-of-concept in our previous work~\cite{hoang_arviz_2021} for providing navigation goals. This method was explored as it allows the chosen goal to be communicated to the user via visualization and AR headsets allow natural gesture detection without additional external setup. The aim of this research is to evaluate this proposed interface against a Tablet interface, which acts similarly to RViz, and person following, which is introduced as a simple and easily understood method. In this research, we envision a scenario where a user is directed to navigate a mobile robot to and from a set number of fixed goals. 

The contribution of this work is an evaluation of a proposed AR Point and Click interface in a user study that aims to investigate the usability of the proposed method against more traditional methods of setting navigation goals to robots: person following and tablet-based interface.

\section{AR Point\&Click Interface}
\vspace{-0.1cm}




The proposed system consists of two main modules, a \textbf{Robot} module which employs ROS and a \textbf{Interface} module which is either an AR headset which uses Unity Engine or an android tablet which employs ROS-Mobile~\cite{rottmann_ros-mobile_2020}. The \textbf{Interface} module communicates the robot's goal to the \textbf{Robot} module, which then uses this information to navigate to a user-specified goal. The \textbf{Robot} module communicates the visualization information to the \textbf{Interface} module. Communication between the modules is achieved through a Wi-Fi router over a local network. We use ROS inter-node communication between the robot and tablet. We employ the ROS\# software package to communicate between the AR headset and robot. Additional information about the environment, such as goals is not displayed or encoded to the \textbf{Interface} module to mimic a free navigation task where goals are not indicated explicitly on the map.

\subsection{Robot Implementation}
\vspace{-0.1cm}
For the \textbf{Robot} module we employ the Fetch Robotics Mobile Manipulator (Fetch robot).
For localization, we use the standard ROS Navigation Stack packages which are modified by Fetch Robotics.
In this study, the robot  navigated between points in the real-world where there are no static obstacles in between them. Therefore, we chose to adopt a simple navigation behavior from our previous works~\cite{gu_seeing_2021,newbury_learning_2020}. Collision avoidance is implemented using the Fetch Robot's LIDAR sensor to detect objects within a certain distance and detects for a roughly 100$\degree$ cone 1 meter in front of the robot. 

\subsection{Augmented Reality Interface}
\vspace{-0.1cm}
We developed an \textbf{AR Point\&Click (AR)} interface. In this interface, the goal is specified with a hand gesture, as implemented by Hoang~\cite{hoang_arviz_2021}. The hand gesture points a ray from the user's arm and and then involves pinching their thumb and index finger. This method is chosen to be an AR implementation of the more standard RViz interface. It employs a Microsoft HoloLens 2 as the AR headset. This headset can display virtual objects on see-through holographic lenses and provides head pose tracking and localizes itself to a fixed reference frame. Using this, a flag pole is visualized at the goal, as shown in Figure~\ref{fig:intro}.

We localize the AR headset with off-the-shelf packages provided by Vuforia~\cite{noauthor_vuforia_nodate} on an origin of the world frame represented by a printed-out AR marker provided by Microsoft. The marker does not need to be within the FOV of the headset but it is possible that the localization drifts. If the localization is off, the user can re-localize the headset by looking at the physical AR marker and uttering the word ``calibrate". To localize the robot and AR headset frames in the same coordinate frame, the AR marker is placed at the corresponding pose of the origin in the Robot's map frame. In this case the AR marker is placed on the floor. Due to errors in localization and marker placement, which are additive, there may be slight mismatches in the frame of the AR headset and Robot.

\section{User Study}
\label{sec:user}
\vspace{-0.1cm}



\begin{figure}[ht]
    \centering
    \vspace{-0.25cm}
    \begin{subfigure}{0.425\linewidth}
    \includegraphics[trim = {0 25 20 11}, clip, width=1\linewidth]{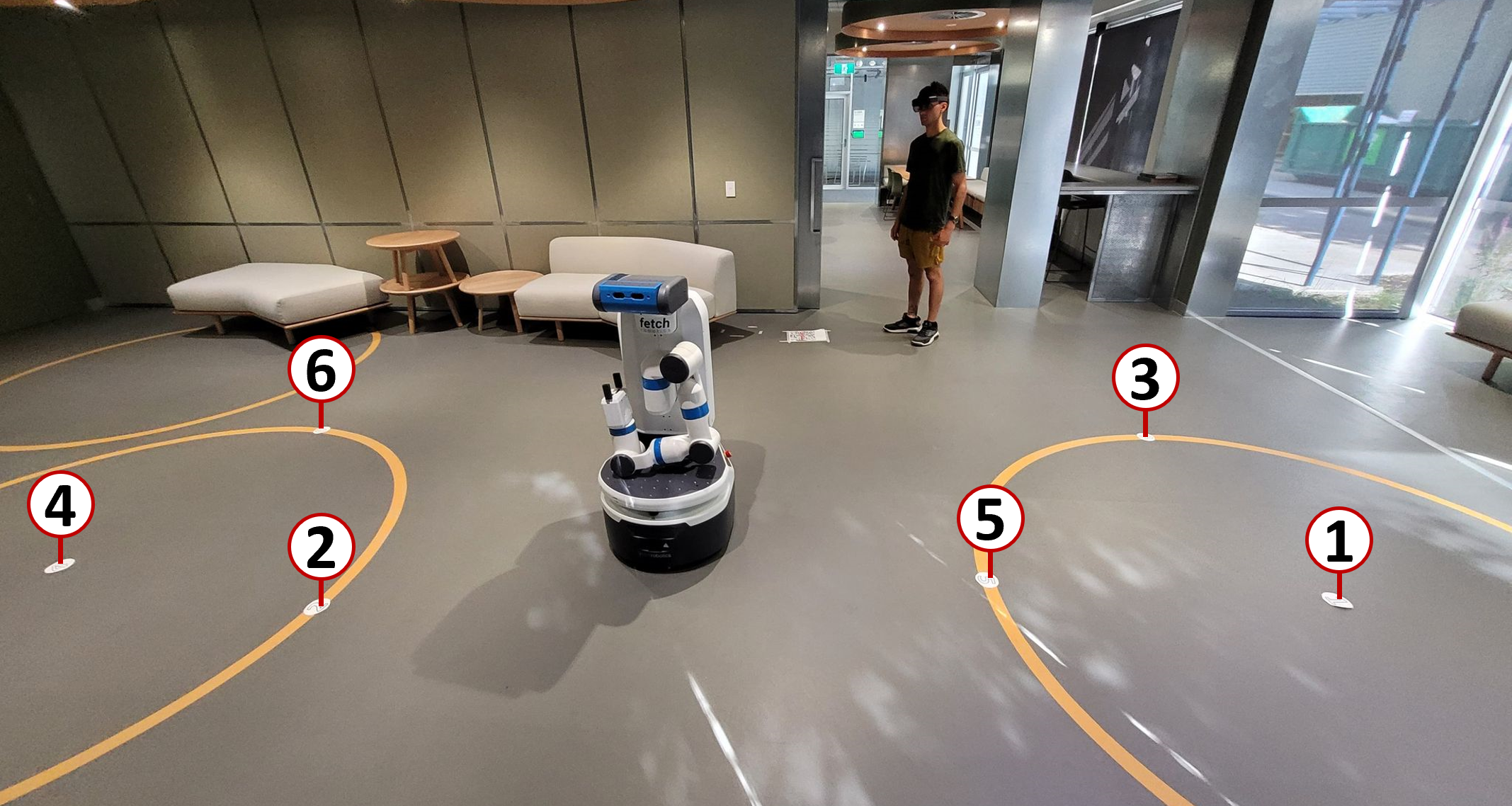}
    \end{subfigure}
    \begin{subfigure}{0.34\linewidth}
    \includegraphics[trim= 30 10 0 7.5, clip, width=1\linewidth]{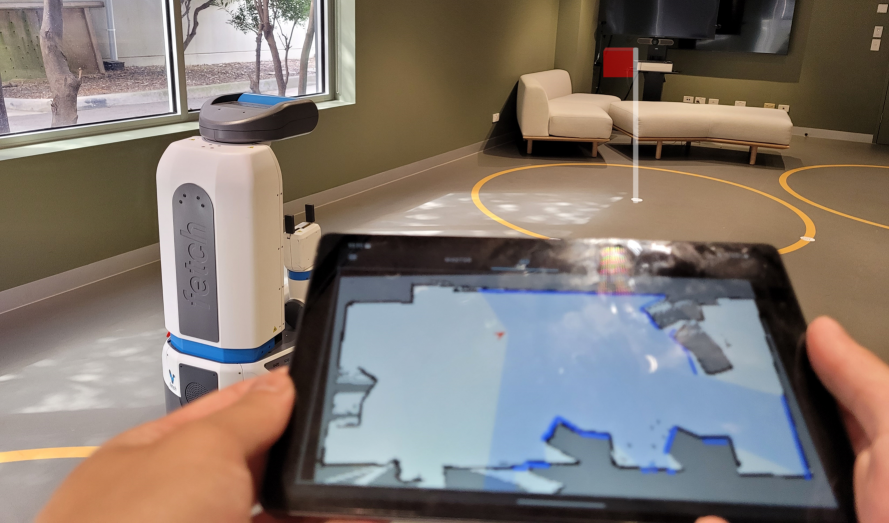}
    \end{subfigure}
    \caption{Left: The initial experimental setup for the user task. The goals are labelled with the numbers. Right: An example of the \textbf{Tablet} interface which uses ROS mobile \cite{rottmann_ros-mobile_2020} that implements the popular RViz interface on mobile devices. Note that the visualized flag is not seen in by the user and is used to symbolize the user's goal.}
    \vspace{-0.6cm}
    \label{fig:study_layout}
\end{figure}

In this study, the user is given the task to navigate the robot to 6 goal positions with a fixed, pre-determined order. The user starts at fixed initial position but can move around the area to navigate the robot. The experimental setup is shown in Figure~\ref{fig:study_layout}. After reaching each goal, the user is asked to wait the robot be become stationary, and then utter the word ``done" to indicate to the experimenter that the goal has been reached. After this occurs, the experimenter specifies the next goal and the user sets the goal position for the robot to navigate. The task is completed when the final goal is reached by the robot. 

In addition to the proposed \textbf{AR Point\&Click} interface, we adopt two baseline interfaces, a \textbf{Person Following} interface and a \textbf{Tablet} interface, which are elaborated on in the section~\ref{sec:baseline}. With these interfaces, our user study has 3 conditions represented by the interfaces. Each user was tested on each condition. The order of the conditions was counterbalanced to reduce ordering effects. For each condition, the user completed a single trial of the navigation task. 

\subsection{Baseline Interfaces}
\vspace{-0.1cm}
\label{sec:baseline}


\textbf{Person Following (PF)} interface: The robot moves directly towards the user within a certain distance. This is activated with a speech command which then uses the pose tracking provided by the HoloLens 2 Headset to obtain the user's position. This interface is designed to be easy to use. 
\vspace{-0.05cm}

\textbf{Tablet} interface: A goal pose is chosen in a top-down 2D map, through ROS Mobile~\cite{rottmann_ros-mobile_2020} by double tapping the screen. The implementation is similar to the standard RViz interface. It visualizes a top-down 2D map, the robot's pose and the laser scanner. Users are only able to rotate or translate the 2D view. 

\vspace{-0.1cm}
\subsection{User Study Procedure}
\label{task}
\vspace{-0.15cm}

The user study was conducted in an open area so that line of sight between the robot and each of the goals could be maintained. Before the user study commenced, the participant read an explanatory statement and filled in a consent form. As well, they watched a prepared video about the task. The experimenter read from a script during the user study to ensure uniformity of procedure.
Each user was given two minutes to practice on each interface before starting the task. This practice time was recorded. Afterwards, the user began the task
on the experimenters signal. After each condition, the user was given a survey to rate the system with questions from Table~\ref{tab:questions}. After completing the experiment, the participants were given the option to engage in a post-experiment interview. The mean duration for each participant was around 30 minutes.

\subsection{Metrics}
\noindent\textbf{Objective metrics:}
\begin{itemize}
    \item Positional Error (m): The Euclidean distance between the desired goal and robot's position when the user deems a navigation goal complete.
    \item Goal-to-goal time (s): The time spent between goals.
    \item Practice time (s): The time from when the participants starts trialing the interface until they decide to stop, or up to 2 minutes maximum.
\end{itemize}

\begin{table}[h!]
\centering
\vspace{-0.55cm}
\caption{User Study Survey Questions}
\begin{tabular}{|l|}
\hline
    Q1: I felt \textbf{safe} during the task.\\
    Q2: I was satisfied with the robot's \textbf{navigation accuracy}.\\
    Q3: The user interface \textbf{aided} my ability to \textbf{complete the task}.\\
    Q4 The interface was \textbf{easy to use}.\\
    Q5 The task was \textbf{mentally demanding}.\\
\hline
\end{tabular}
\label{tab:questions}
\vspace{-0.25cm}
\end{table}

\noindent\textbf{Subjective metrics:} After each task is completed, the participant was asked to respond to a set of questions, shown in Table~\ref{tab:questions}. The questions were all measured on a 7-point Likert scale and were designed to focus on the perceived usability and efficacy of the system. After the participant completed the task on all three interfaces, they are asked for their preferred user interface for this task.

\subsection{Hypotheses}

\vspace{-1cm}
\begin{table}[h]
\centering
\caption{Hypotheses}
\begin{tabular}{|l l|}
\hline
    \textbf{H1:} &  The \textbf{AR Point\&Click} method will have the highest \textbf{Perceived Safety}.\\\hline
    \textbf{H2:} & The \textbf{AR Point\&Click} and \textbf{Person Following} interfaces will be rated \\
    & significantly higher in \textbf{Perceived Accuracy} than the \textbf{Tablet} interface.\\\hline
    \textbf{H3:} &  The \textbf{AR Point\&Click} and \textbf{Person Following} interfaces will be rated \\
    & significantly higher in \textbf{Perceived Task Efficiency} than the \textbf{Tablet} interface.\\\hline
    \textbf{H4:} &  The \textbf{Person Following} interface will be the \textbf{Easiest to Use}.\\\hline
    \textbf{H5:} &  The \textbf{Tablet} interface will be the most \textbf{Mentally Demanding} to use.\\\hline
    \textbf{H6:} &  The \textbf{AR Point\&Click} interface will have the lowest \textbf{Positional Error} \\
    & compared to other methods\\
\hline
\end{tabular}
\vspace{-0.2cm}
\label{tab:hypotheses}
\end{table}

We expect that the AR based interface will result in improved positional accuracy and perceived safety. We also expect reduced mental load and improved perception of task efficiency and accuracy for both the person following and AR based interfaces. As well, we predict that the Person Following interface will be most intuitive. As such we formulate the following hypotheses in Table~\ref{tab:hypotheses}.



\section{Quantitative Results}
For this study we recruited 18 participants from within the university\footnote[1]{Due to the COVID-19 pandemic, no external participants could be recruited. This study has been approved by the Monash University Human Research Ethics Committee (Application ID: 27685)}, including 16 male and 2 female participants, between the ages of 21 and 33 ($\mu=24.4$, $\sigma=3.6$). All but one of the participants had prior experience in robotics. 6 of the 54 trials were repeated due to AR application crashes. Prior to any task employing AR applications, user's were informed that they would repeat the task in the case of an application crash. They were informed that this error should not be reflected in their responses to the survey questions.

\begin{figure*}[ht]
\centering
\vspace{-0.35cm}
\begin{tikzpicture}[scale=1]
    \definecolor{r}{RGB}{255,0,32}
    \definecolor{rw1}{RGB}{230, 50, 60}
    \definecolor{rw2}{RGB}{213, 123, 111}
    \definecolor{w}{RGB}{198, 198, 198}
    \definecolor{bw2}{RGB}{142, 160, 188}
    \definecolor{bw1}{RGB}{60, 130, 200}
    \definecolor{b}{RGB}{5,113,200}

    \pgfplotsset{
       /pgfplots/bar  cycle  list/.style={/pgfplots/cycle  list={%
            {r!75!black,fill=r!60!white,mark=none},%
            {rw1!75!black,fill=rw1!60!white,mark=none},%
            {rw2!75!black,fill=rw2!60!white,mark=none},%
            {w!75!black,fill=w!60!white,mark=none},%
            {bw2!75!black,fill=bw2!60!white,mark=none},%
            {bw1!75!black,fill=bw1!60!white,mark=none},%
            {b!75!black,fill=b!60!white,mark=none},%
         }
       },
    }
    
    \tikzstyle{every node}=[font=\small]

    \begin{axis}[
        name=mainplot,
        xbar stacked,
        title=\textbf{Perceived Safety},
        nodes near coords,
        bar width=0.8,
        width = 0.4\textwidth,
        height = 0.25\textwidth,
        xmin = 0, xmax = 18,
        enlarge y limits={abs=6pt},
        ytick={0,1,2},
        yticklabels={T, PF, AR},  
        xtick={0,7.5,15},
        xticklabels={0\%,50\%,100\%},         
        legend style={at={(0.5,-0.20)}, anchor=north, legend columns=-1, /tikz/every even column/.append style={column sep=0.5cm}},
    ]
    
        \addplot coordinates
        {(0,2) (0,1) (0,0)};
        
        \addplot coordinates
        {(0,2) (0,1) (0,0)};
        
        \addplot coordinates
        {(0,2) (0,1) (0,0)};
        
        \addplot coordinates
        {(0,2) (0,1) (4,0)};
        
        \addplot coordinates
        {(0,2) (5,1) (2,0)};
        
        \addplot coordinates
        {(7,2) (5,1) (5,0)};  
        
        \addplot coordinates
        {(11,2) (8,1) (7,0)};    
        
    \end{axis}
    \begin{axis}[
        name=secondplot,
        title=\textbf{Perceived Accuracy},
        at={(mainplot.north east)},
        xshift=0.75cm,
        anchor=north west,    
        xbar stacked,
        nodes near coords,
        bar width=0.8,
        width = 0.4\textwidth,
        height = 0.25\textwidth,
        xmin = 0, xmax = 18,
        enlarge y limits={abs=6pt},
        ytick={0,1,2},
        yticklabels={,,,},  
        xtick={0,7.5,15},
        xticklabels={0\%,50\%,100\%},         
        legend style={at={(0.5,-0.20)}, anchor=north, legend columns=-1, /tikz/every even column/.append style={column sep=0.5cm}},
    ]
    
        \addplot coordinates
        {(0,2) (0,1) (0,0)};
        
        \addplot coordinates
        {(0,2) (0,1) (3,0)};
        
        \addplot coordinates
        {(0,2) (0,1) (2,0)};
        
        \addplot coordinates
        {(1,2) (1,1) (4,0)};
        
        \addplot coordinates
        {(1,2) (1,1) (3,0)};
        
        \addplot coordinates
        {(5,2) (9,1) (3,0)};  
        
        \addplot coordinates
        {(11,2) (7,1) (3,0)};
        
    \end{axis}
    \begin{axis}[
        name=thirdplot,
        title=\textbf{Perceived Efficiency},
        at={(secondplot.north east)},
        xshift=0.75cm,
        anchor=north west,      
        xbar stacked,
        nodes near coords,
        bar width=0.8,
        width = 0.4\textwidth,
        height = 0.25\textwidth,
        xmin = 0, xmax = 18,
        enlarge y limits={abs=6pt},
        ytick={0,1,2.5,3.5},
        ytick={0,1,2},
        yticklabels={,,,},  
        xtick={0,7.5,15},
        xticklabels={0\%,50\%,100\%},         
        legend style={at={(0.5,-0.20)}, anchor=north, legend columns=-1, /tikz/every even column/.append style={column sep=0.5cm}},
    ]
    
        \addplot coordinates
        {(0,2) (0,1) (0,0)};
        
        \addplot coordinates
        {(0,2) (0,1) (2,0)};
        
        \addplot coordinates
        {(0,2) (0,1) (4,0)};
        
        \addplot coordinates
        {(1,2) (1,1) (2,0)};
        
        \addplot coordinates
        {(2,2) (1,1) (6,0)};
        
        \addplot coordinates
        {(7,2) (9,1) (3,0)};  
        
        \addplot coordinates
        {(8,2) (7,1) (1,0)};    
        
    \end{axis}
    \begin{axis}[
        at={(mainplot.below south east)},
        title=\textbf{Ease of Use},
        yshift=-0.6cm,
        anchor=north,
        xbar stacked,
        nodes near coords,
        bar width=0.8,
        width = 0.4\textwidth,
        height = 0.25\textwidth,
        xmin = 0, xmax = 18,
        enlarge y limits={abs=6pt},
        ytick={0,1,2},
        yticklabels={T, PF, AR},  
        xtick={0,7.5,15},
        xticklabels={0\%,50\%,100\%},         
        legend style={at={(0.5,-0.20)}, anchor=north, legend columns=-1, /tikz/every even column/.append style={column sep=0.5cm}},
    ]
    
        \addplot coordinates
        {(0,2) (0,1) (0,0)};
        
        \addplot coordinates
        {(0,2) (0,1) (4,0)};
        
        \addplot coordinates
        {(2,2) (0,1) (3,0)};
        
        \addplot coordinates
        {(1,2) (0,1) (2,0)};
        
        \addplot coordinates
        {(4,2) (2,1) (2,0)};
        
        \addplot coordinates
        {(7,2) (6,1) (3,0)};  
        
        \addplot coordinates
        {(4,2) (10,1) (4,0)};     
        
    \end{axis}
    \begin{axis}[
        title=\textbf{Mental Load} (Reverse Scale),
        at={(secondplot.below south east)},
        yshift=-0.6cm,
        anchor=north,    
        xbar stacked,
        nodes near coords,
        bar width=0.8,
        width = 0.4\textwidth,
        height = 0.25\textwidth,
        xmin = 0, xmax = 18,
        enlarge y limits={abs=6pt},
        ytick={0,1,2},
        yticklabels={,,,},  
        xtick={0,7.5,15},
        xticklabels={0\%,50\%,100\%},         
        legend style={at={(0,-0.40)}, anchor=north , legend columns=-1, /tikz/every even column/.append style={column sep=0.5cm}},
    ]
    
        \addplot coordinates
        {(6,2) (6,1) (0,0)};
        
        \addplot coordinates
        {(4,2) (7,1) (3,0)};
        
        \addplot coordinates
        {(5,2) (1,1) (4,0)};
        
        \addplot coordinates
        {(2,2) (1,1) (1,0)};
        
        \addplot coordinates
        {(1,2) (1,1) (5,0)};
        
        \addplot coordinates
        {(0,2) (2,1) (4,0)};  
        
        \addplot coordinates
        {(0,2) (0,1) (1,0)};       
        
        \legend{St.D, D, Sl.D, N, Sl.A, A, St.A} 
        
    \end{axis}

\end{tikzpicture}
\vspace{-0.35cm}
\caption{Summary of the raw data obtained from the survey filled out by $18$ participants, comparing the \textbf{AR Point\&Click} (AR), \textbf{Person Following} (PF) and \textbf{Tablet} (T) interfaces, for each one of the 5 survey questions shown in Table~\ref{tab:questions}. (St = Strongly, Sl = Slightly, D = Disagree, N = Neutral, A = Agree)}
\label{fig:barplot}
\vspace{-0.2cm}
\end{figure*}
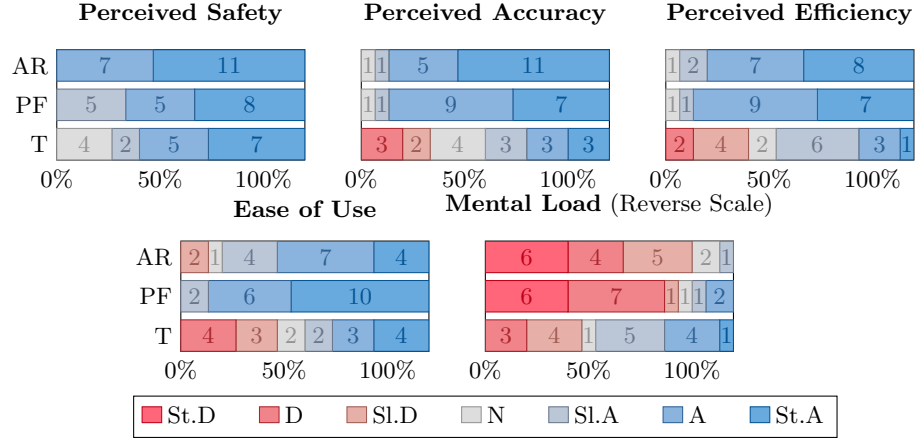

\subsection{Subjective Metrics}

\renewcommand{\arraystretch}{0.7}
\begin{table}[ht]
\small
\vspace{-0.75cm}
\caption{Post-hoc Nemenyi test results for the survey questions used to test \textbf{H1}-\textbf{H5}. Significant results are indicated in \textbf{bold} and was determined for p-values $\textless$ 0.05.}
\begin{tabular*}{\columnwidth}{@{}l@{\extracolsep{\fill}}ccccc@{}}
\toprule
& $\bm{p}$ (AR $\neq$ PF) &  $\bm{p}$ (AR $\textgreater$ T) & $\bm{p}$ (PF $\textgreater$ T)\\ 
\toprule
Perceived Safety & 0.29 & 0.06 & 0.73 \\
Perceived Accuracy & 0.83 & \textbf{$<$0.001} & \textbf{0.005} \\
Perceived Efficiency & 0.99 & \textbf{0.002} & \textbf{0.001} \\
Ease of Use & 0.077 & 0.63 & \textbf{0.005}\\
Mental Load & 0.99 &\textbf{0.001} & \textbf{$<$0.001}\\\bottomrule
\vspace{-0.2cm}
\end{tabular*}
\label{table:nemenyi}
\end{table}

The raw response distribution for each survey question is shown in Fig.~\ref{fig:barplot}.
To test \textbf{H1-H5}, we employed a Friedman test on the likert scale variables as the ordinal nature these require a non-parametric test for more than 2 categories. Significance was found in all questions and a post-hoc Nemenyi test was then conducted between interfaces with p-values shown in Table~\ref{table:nemenyi}.


Regarding \textit{Perceived Safety} (Q1), no statistical significance between any two methods were found, therefore \textbf{H1} cannot be affirmed. We do, however, observe a trend that the \textbf{AR Point\&Click} is rated highest by the participants, with 11 people strongly agreeing that they felt safe. We believe that a larger participant pool might affirm this hypothesis, especially for the hypothesis that \textbf{AR Point\&Click} would be found significantly safer than the \textbf{Tablet} method (p=0.06, which is close to the significance boundary). We also observe that the \textit{Perceived Safety} (Q1) question was rated the highest overall among all 5 questions, signalling that the participants felt safe interacting with the robot, which is an important consideration for any human-robot interaction scenario.

Participants rated the \textbf{Person Following} and \textbf{AR Point\&Click} interfaces significantly higher than the \textbf{Tablet} interface in terms of \textit{Perceived Accuracy} (Q2) and \textit{Perceived Task Efficiency} (Q3), affirming \textbf{H2} and \textbf{H3}, respectively. Moreover, both of these methods were rated to require significantly less \textit{Mental Load }(Q5) than the \textbf{Tablet} interface, affirming \textbf{H5}. No difference between \textbf{AR Point\&Click} and \textbf{Person Following} interfaces was found for these metrics.
The \textbf{Person Following} method was found to be the most \textit{Easy to Use (Q4)} with significant difference found with the \textbf{Tablet} (p=0.005). However signifiance was not found with the \textbf{AR Point\&Click} interface (p=0.077). 
Therefore, \textbf{H4} can only be partially affirmed and further user studies and feedback is warranted.

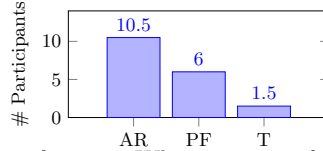
\begin{figure}[ht]
    \centering
    \vspace{-0.35cm}
    \begin{tikzpicture}[scale=0.8]
    \begin{axis}[
        ybar,
        enlarge x limits=0.5,
        legend style={at={(0.5,-0.25)},
        anchor=north,legend columns=-1,
        /tikz/every even column/.append style={column sep=0.5cm}},
        ylabel={\# Participants},
        symbolic x coords={AR, PF, T},
        xtick=data,
        nodes near coords,
        nodes near coords align={vertical},
        bar width=25pt,
        xtick pos=left,
        width = 0.485\textwidth,
        height = 0.275\textwidth,
        ymin = 0, ymax = 14,
        legend image code/.code={\draw [#1] (0cm,-0.075cm) rectangle (0.2cm,0.25cm); },
    ]
    
    \addplot coordinates {(AR,10.5) (PF,6) (T,1.5)};
    
    
    \end{axis}
\end{tikzpicture}
    \vspace{-0.15cm}
    \caption{Distribution of user preferences. Where users preferred more than one interface, fractions were used. The AR Point\&Click was the most preferred interface}
    \label{fig:preferences}
    \vspace{-0.4cm}
\end{figure}

When asked for interface preference for this task after being exposed to all three interfaces, 12 users out of 18 indicated that they prefer the \textbf{AR Point\&Click} interface, including 3 who indicated that they equally preferred another method. The distribution of user preferences are shown in Figure~\ref{fig:preferences}.

\vspace{-0.1cm}

\subsection{Objective Metrics}
\vspace{-0.2cm}
The mean and standard deviation for the three objective metrics and interfaces are shown in Table~\ref{table:objResults}. While \textbf{Person Following} had the lowest Practice Time and Goal Time, \textbf{AR Point\&Click} had the lowest Positional Error.

\begin{table}[h]
\vspace{-0.55cm}
\caption{Mean and standard deviation for each objective metric and interface. Lower numbers are better - and the best performing methods are indicated in \textbf{bold}.}
\centering
\setlength{\tabcolsep}{2.8pt}
\begin{tabular}{lccc}
\toprule
 & AR & PF & T   \\ 
\toprule
Practice time (s) & 84.1 ($\sigma$ = 40.1) & \textbf{61.4 ($\sigma$ = 26.2)} & 74.7 ($\sigma$ = 30.8)\\
Goal Time (s) & 19.3 ($\sigma$ = 7.3) & \textbf{16.8 ($\sigma$ = 3.7)} & 24.5 ($\sigma$ = 4.9) \\
Pos. Error (m) & \textbf{0.137 ($\sigma$ = 0.03) }& 0.141 ($\sigma$ = 0.04) & 0.188 ($\sigma$ = 0.05) \\\bottomrule
\end{tabular}
\vspace{-0.25cm}
\label{table:objResults}
\end{table}

We conducted a one-way ANOVA test in which we found that there was no statistical significance in the Practice Time metric. There were statistically significant differences for the Goal Time and Positional Error metrics which we subsequently ran post-hoc t-tests for. Bonferroni method~\cite{pvalueadjust} was employed to account for the repeated tests, which reduced the threshold for statistical significance by a factor of 3, hence we used p=0.0167 as the statistical significance threshold. The t-test results for each objective metric pair are shown in Table~\ref{table:ttest}. 

\begin{table}[ht]
\small
\vspace{-0.55cm}
\caption{p-values for post-hoc t-test results for the objective metrics. Significant results and preferred method are indicated in \textbf{bold}.}
\begin{tabular*}{\columnwidth}{@{}l@{\extracolsep{\fill}}ccccc@{}}
\toprule
& $\bm{p}$ (AR $\neq$ PF) &  $\bm{p}$ (AR $\textgreater$ T) & $\bm{p}$ (PF $\textgreater$ T)\\ 
\toprule
Goal Time & 0.25 & 0.030 & \textbf{$<$0.001} \\
Positional Error & 0.74 & \textbf{0.001} & \textbf{$<$0.001} \\\bottomrule
\end{tabular*}
\vspace{-0.3cm}
\label{table:ttest}
\end{table}

While we found that \textbf{AR Point\&Click} interface resulted in the lowest Positional Error, and it was significantly lower than \textbf{Tablet} interface, we could not find a statistical difference between \textbf{AR Point\&Click} and \textbf{Person Following}. Hence, \textbf{H6} can not be affirmed. Overall, the user study results supports the hypotheses \textbf{H2}, \textbf{H3}, and \textbf{H5} and partially supports  \textbf{H4}, but does not have statistically significant results to affirm \textbf{H1} and \textbf{H6}.  
\vspace{-0.05cm}
\section{Qualitative Analysis}
\vspace{-0.15cm}

During the user studies, we noticed that the participants' perception of accuracy and mental load appeared to be linked to ease of fine adjustments in position rather than the actual accuracy of the robot's position. When using the \textbf{Tablet} interface, one participant (R16) found that they \textit{"had to make many adjustments"}. Meanwhile another participant (R17) stated that \textit{"it was nice to be able to see where you were going to send the robot"} in reference to the \textbf{AR Point\&Click} interface. 
Unexpectedly, we observed that users made fine adjustments with the \textbf{Person Following} interface by making small movements for the robot to follow. This, along with comments about the \textbf{AR Point\&Click} interface, suggests that positional accuracy and perceived efficiency improve with the ability to localize goals within the physical space. A participant (R12) suggested that \textit{"the best interface would be a combination of the }\textbf{[AR Point\&Click]}\textit{ and }[\textbf{Person Following}] \textit{interfaces"}. 
It can be noted that the \textbf{AR Point\&Click} had similar improvements over the \textbf{Tablet} interface as the \textbf{Person Following} interface with the exception of \textbf{Q4}, where users did not find the AR interface easy to use relative to the others. One participant (R5) notes that the \textbf{AR Point\&Click} interface took \textit{"longer to figure out but once [they] figured it out, it was quite good to use"}. This suggests that the short calibration time of the user study may not be fully representative of or sufficient for the interface. 

Though \textbf{H1} was not affirmed, the perceived safety of the \textbf{AR Point\&Click} may be a key factor in its adoption as the AR headset allowed users to look at the robot instead of away at a \textbf{Tablet} or being in close proximity to the robot in the \textbf{Person Following} interface. A participant (R11) found that \textit{"you can observe the motion of the robot"} with the \textbf{AR Point\&Click} interface while also stating that \textit{"It was a little bit scary"} when using the \textbf{Person Following} interface.
An interesting observation was that most of the participants expressed a preference for the \textbf{AR Point\&Click} interface, even though it did not perform significantly better than \textbf{Person Following} in any metric. As well, the participants are of similar ages and all but one had prior experience in robotics. These results might be a novelty effect of AR technology, which motivates long-term studies. 



\vspace{-0.1cm}
\section{Conclusions and Future Work}
\vspace{-0.15cm}
We investigate a point and click interface using an Augmented Reality headset which allows users to employ a natural point-and-click gesture to set navigation goals. The interface also visualizes the goal after it was set which lets users adjust the goal location if desired. We validate the efficacy of the proposed interface through a user study, in which participants set navigation goals for a robot. We compare the proposed \textbf{AR Point\&Click} interface to a \textbf{Tablet} interface which is similar to the popular developer tool RViz, and a \textbf{Person Following} behavior, designed to be an intuitive interface. 

User study results show that the users set goals with the highest accuracy using the proposed \textbf{AR Point\&Click} interface. Moreover, \textbf{AR Point\&Click} was subjectively assessed to have higher perceived accuracy and efficiency and a lower mental load compared to the \textbf{Tablet} interface. Though not conclusive, there was some indication that the \textbf{AR Point\&Click} interface also improved the perceived safety. The proposed interface was rated similarly to \textbf{Person Following} by the participants, however, the majority of the participants indicated that \textbf{AR Point\&Click} as their interface of choice, likely due to a novelty effect. On the other hand, there was some indication that \textbf{Person Following} method was found to be easier to use than \textbf{AR Point\&Click}. This may be alleviated with greater adoption of the AR technology in the future. Given these results, the proposed \textbf{AR Point\&Click} is a feasible method for setting navigation goals for mobile robots and there is room for improvement in the user experience.

Some of the main drawbacks found for the \textbf{AR Point\&Click} were due to limitations of the AR headset. Participants found that (R11) \textit{"[you] couldn't exactly look at the point"} due to the AR headset's low FOV and often noticed that the click gesture recognition wouldn't register or would falsely register. It is important to note that the improvements to the AR technology in the future can alleviate some of the issues experienced by users.
It should be noted that this work is a proof-of-concept for using an AR headset to control mobile robots and there are several ways to improve on this. First, the feasibility of the navigation goals are not checked, hence collision checks can be incorporated. Second, the user study can be extended to test how well people can set goal poses instead of just position goals. Third, the current navigation behavior is very simple and therefore implementing a more complex navigation algorithm, which may be human-aware \cite{cosgun2016anticipatory}, may be another avenue to investigate. Finally, the proposed system can be extended to select other entities, such as people or objects.
\vspace{-0.1cm}
\section{Acknowledgement}
\vspace{-0.15cm}
This project was supported by the Australian Research Council (ARC) Discovery Project Grant DP200102858.

\balance

\bibliographystyle{splncs04}
\bibliography{references.bib}

\begin{thebibliography}{10}
\providecommand{\url}[1]{\texttt{#1}}
\providecommand{\urlprefix}{URL }
\providecommand{\doi}[1]{https://doi.org/#1}

\bibitem{service_robots}
{USD 35.27 Billion Growth in Service Robotics Market: By Application
  (professional robots and personal robots) and Geography - Global Forecast to
  2025}.
  \url{https://www.prnewswire.com/news-releases/usd-35-27-billion-growth-in-service-robotics-market-by-application-professional-robots-and-personal-robots-and-geography---global-forecast-to-2025--301454915.html},
  2022-03-24

\bibitem{noauthor_vuforia_nodate}
{Vuforia} {Developer} {Portal} {\textbar}, \url{https://developer.vuforia.com/}

\bibitem{chen2013context}
Chen, Y.H., Zhang, B., Tuna, C., Li, Y., Lee, E.A., Hartmann, B.: A context
  menu for the real world: Controlling physical appliances through head-worn
  infrared targeting. Tech. rep., UC Berkeley EECS (2013)

\bibitem{cosgun_context-aware_2018}
Cosgun, A., Christensen, H.I.: Context-aware robot navigation using
  interactively built semantic maps. Paladyn, Journal of Behavioral Robotics
  \textbf{9}(1) (2018)

\bibitem{cosgun_autonomous_2013}
Cosgun, A., Florencio, D.A., Christensen, H.I.: Autonomous person following for
  telepresence robots. In: {IEEE} {International} {Conference} on {Robotics}
  and {Automation} (ICRA) (2013)

\bibitem{cosgun2015human}
Cosgun, A., Maliki, A., Demir, K., Christensen, H.: Human-centric assistive
  remote control for co-located mobile robots. In: ACM/IEEE International
  Conference on Human-Robot Interaction (HRI) Extended Abstracts. pp. 27--28
  (2015)

\bibitem{cosgun2016anticipatory}
Cosgun, A., Sisbot, E.A., Christensen, H.I.: Anticipatory robot path planning
  in human environments. In: IEEE international symposium on robot and human
  interactive communication (RO-MAN) (2016)

\bibitem{cosgun2015did}
Cosgun, A., Trevor, A.J., Christensen, H.I.: Did you mean this object?:
  detecting ambiguity in pointing gesture targets. In: HRI’15 Towards a
  Framework for Joint Action Workshop (2015)

\bibitem{gu_seeing_2021}
Gu, M., Cosgun, A., Chan, W.P., Drummond, T., Croft, E.: Seeing thru walls:
  Visualizing mobile robots in augmented reality. In: IEEE International
  Conference on Robot and Human Interactive Communication (RO-MAN) (2021)

\bibitem{gualtieri2017open}
Gualtieri, M., Kuczynski, J., Shultz, A.M., Ten~Pas, A., Platt, R., Yanco, H.:
  Open world assistive grasping using laser selection. In: IEEE International
  Conference on Robotics and Automation (ICRA) (2017)

\bibitem{hato2010pointing}
Hato, Y., Satake, S., Kanda, T., Imai, M., Hagita, N.: Pointing to space:
  modeling of deictic interaction referring to regions. In: ACM/IEEE
  International Conference on Human-Robot Interaction (HRI) (2010)

\bibitem{hoang_virtual_2021}
Hoang, K.C., Chan, W.P., Lay, S., Cosgun, A., Croft, E.: Virtual {Barriers} in
  {Augmented} {Reality} for {Safe} and {Effective} {Human}-{Robot}
  {Cooperation} in {Manufacturing}. In: IEEE International Conference on Robot
  and Human Interactive Communication (RO-MAN) (2022)

\bibitem{newbury_visualizing_2021}
Hoang, K.C., Chan, W.P., Lay, S., Cosgun, A., Croft, E.: Visualizing {Robot}
  {Intent} for {Object} {Handovers} with {Augmented} {Reality}. In: IEEE
  International Conference on Robot and Human Interactive Communication
  (RO-MAN) (2022)

\bibitem{hoang_arviz_2021}
Hoang, K.C., Chan, W.P., Lay, S., Cosgun, A., Croft, E.: Arviz: An augmented
  reality-enabled visualization platform for ros applications. IEEE Robotics
  Automation Magazine pp. 2--11 (2022)

\bibitem{pvalueadjust}
Jafari, M., Ansari-Pour, N.: Why, when and how to adjust your p values? Cell
  journal  \textbf{20},  604--607 (2019)

\bibitem{kemp2008point}
Kemp, C.C., Anderson, C.D., Nguyen, H., Trevor, A.J., Xu, Z.: A point-and-click
  interface for the real world: laser designation of objects for mobile
  manipulation. In: ACM/IEEE International Conference on Human-Robot
  Interaction (HRI) (2008)

\bibitem{kousi_enabling_2019}
Kousi, N., Stoubos, C., Gkournelos, C., Michalos, G., Makris, S.: Enabling
  {Human} {Robot} {Interaction} in flexible robotic assembly lines: an
  {Augmented} {Reality} based software suite. Procedia CIRP  (2019)

\bibitem{makhataeva2020augmented}
Makhataeva, Z., Varol, H.A.: Augmented reality for robotics: A review. Robotics
   \textbf{9}(2), ~21 (2020)

\bibitem{newbury_learning_2020}
Newbury, R., Cosgun, A., Koseoglu, M., Drummond, T.: Learning to {Take} {Good}
  {Pictures} of {People} with a {Robot} {Photographer}. In: {IEEE}/{RSJ}
  {International} {Conference} on {Intelligent} {Robots} and {Systems} ({IROS})
  (2020)

\bibitem{nguyen2008clickable}
Nguyen, H., Jain, A., Anderson, C., Kemp, C.C.: A clickable world: Behavior
  selection through pointing and context for mobile manipulation. In: IEEE/RSJ
  International Conference on Intelligent Robots and Systems (IROS) (2008)

\bibitem{reardon_come_2018}
Reardon, C., Lee, K., Fink, J.: Come {See} {This}! {Augmented} {Reality} to
  {Enable} {Human}-{Robot} {Cooperative} {Search}. In: 2018 {IEEE}
  {International} {Symposium} on {Safety}, {Security}, and {Rescue} {Robotics}
  ({SSRR}) (2018)

\bibitem{rottmann_ros-mobile_2020}
Rottmann, N., Studt, N., Ernst, F., Rueckert, E.: {ROS}-{Mobile}: {An}
  {Android} application for the {Robot} {Operating} {System}. arXiv preprint
  arXiv:2011.02781  (2020)

\bibitem{scales_studying_2020}
Scales, P., Aycard, O., Aubergé, V.: Studying {Navigation} as a {Form} of
  {Interaction}: a {Design} {Approach} for {Social} {Robot} {Navigation}
  {Methods}. In: {IEEE} {International} {Conference} on {Robotics} and
  {Automation} ({ICRA}) (2020)

\bibitem{sprute2019far}
Sprute, D., T{\"o}nnies, K., K{\"o}nig, M.: This far, no further: Introducing
  virtual borders to mobile robots using a laser pointer. In: IEEE
  International Conference on Robotic Computing (IRC). pp. 403--408 (2019)

\bibitem{walker_communicating_2018}
Walker, M., Hedayati, H., Lee, J., Szafir, D.: Communicating robot motion
  intent with augmented reality. In: ACM/IEEE International Conference on
  Human-Robot Interaction (HRI) (2018)

\bibitem{waymouth_demonstrating_2021}
Waymouth, B., Cosgun, A., Newbury, R., Tran, T., Chan, W.P., Drummond, T.,
  Croft, E.: Demonstrating cloth folding to robots: Design and evaluation of a
  2d and a 3d user interface. In: IEEE International Conference on Robot and
  Human Interactive Communication (RO-MAN) (2021)

\end{thebibliography}
\end{document}